\documentclass{ecai} 

\usepackage{latexsym}
\usepackage{amssymb}
\usepackage{amsmath}
\usepackage{amsthm}
\usepackage{graphicx}
\usepackage{xspace}
\graphicspath{{figures/}}
\usepackage[table]{xcolor} 
\usepackage{colortbl} 
\usepackage{booktabs}

\newcommand{\term}[1]{\textit{#1}}
\newcommand{\quotes}[1]{`#1'}
\newcommand{\qpvt}{$Q_\textrm{PVT}$\xspace}

\begin{document}

\begin{frontmatter}

\title{Predicting Solar Heat Production \\
 to Optimize Renewable Energy Usage}

\author[A]{\fnms{Tatiana}~\snm{Boura}\orcid{0009-0008-0656-4372}\thanks{Corresponding Author. Email: tatianabou@iit.demokritos.gr.}}
\author[A]{\fnms{Natalia}~\snm{Koliou}\orcid{0009-0004-3920-9992}}
\author[B]{\fnms{George}~\snm{Meramveliotakis}\orcid{0000-0002-0110-347X}}
\author[A]{\fnms{Stasinos}~\snm{Konstantopoulos}\orcid{0000-0002-2586-1726}}
\author[B]{\fnms{George}~\snm{Kosmadakis}\orcid{0000-0002-3671-8693}}

\address[A]{Institute of Informatics and Telecommunications,
  NCSR~\quotes{Demokritos}, Ag.~Paraskevi, Greece}
\address[B]{Institute of Nuclear \& Radiological Sciences and Technology, Energy \& Safety,
  NCSR~\quotes{Demokritos}, Ag.~Paraskevi, Greece}

\begin{abstract}
Utilizing solar energy to meet space heating and domestic hot water
demand is very efficient (in terms of environmental footprint as well
as cost), but in order to ensure that user demand is entirely covered
throughout the year needs to be complemented with auxiliary heating
systems, typically boilers and heat pumps. Naturally, the optimal
control of such a system depends on an accurate prediction of solar
thermal production.

Experimental testing and physics-based numerical models are used to
find a collector's performance curve --- the mapping from solar
radiation and other external conditions to heat production --- but this
curve changes over time
once the collector is exposed to outdoor conditions. In order to
deploy advanced control strategies in small domestic installations,
we present an approach that uses machine learning to automatically
construct and continuously adapt a model that predicts heat
production. Our design is driven by the need to
(a) construct and adapt models using supervision that can be extracted
from low-cost instrumentation, avoiding extreme accuracy and
reliability requirements; and
(b) at inference time, use inputs that are typically provided in
publicly available weather forecasts.

Recent developments in attention-based machine learning, as well
as careful adaptation of the training setup to the specifics of the task,
have allowed us to design a machine learning-based solution that
covers our requirements.
We present positive empirical results for the predictive accuracy
of our solution, and discuss the impact of these results on the
end-to-end system.
\end{abstract}

\end{frontmatter}

\hfill \break

\section{Introduction}

Efficient utilization of renewable energy, especially solar
energy through solar thermal or PV/thermal (PVT) collectors, is
a promising solution to meet the
needs for heating and domestic hot water, offering substantial
advantages with respect to both environmental footprint and operational
costs. Particularly during mid-seasons and summer, the available solar
radiation effectively eliminates the need for conventional boilers.
However, to ensure that user demand is entirely covered
throughout the year, collectors are complemented with auxiliary
heating systems, typically boilers and heat pumps. 

The controller of such a system should minimize the use of
auxiliary heating by employing insulated hot water tanks to shift
the load from when demand is high to when there is excess production.
Although this is relatively straight-forward for space heating,
combined space heating and domestic hot water usage is more
challenging as the latter experiences demand spikes.
Individual usage patterns can be user-parameterized or statistically
estimated, but optimal
control also requires reliable prediction of solar thermal
production to match the demand and reduce heat losses from the
storage tank.

Given the importance of estimating solar thermal collectors'
performance, it is only natural that the problem has been 
approached from multiple angles. Based on experimental testing,
the methods for estimating the efficiency of collectors
\emph{under specified conditions} have been standardized
(ASHRAE 93, ISO~9806, EN12976-2). These methods provide the
parameters required to predict the long term performance of
solar thermal collectors \citep{rojas-etal:2008}.
Numerical modelling methods have also been developed to
estimate the heat production and efficiency of collectors
\citep{cadafalch:2008,khelifa-etal:2016}.
These models account for heat transfer mechanisms such as conduction,
convection, and radiation under known plate geometry and operating
conditions and serve as useful tools for the theoretical
design and optimization of these components.

Accurate as they might initially be, the major shortcoming of
experimental and numerical methods is that the performance curve
of the collector changes over time once the collector is exposed
to outdoor conditions. In order to deploy advanced control strategies
at scale, we need an approach that does not rely on prior knowledge
of the performance curve, but can automatically construct and
continuously adapt a model that predicts heat production based on
expected external conditions (solar radiation and ambient temperature).
Machine learning is a natural fit for such a task, and
unsurprisingly, there is a rich literature on applying machine
learning to construct predictors for thermal collectors'
heat output.

In the work presented here, we focus on the additional requirements
for the mass-deployment of advanced control strategies in
small domestic installations. Besides the continuous adaptation
covered by using machine learning in general (presented in
Section~\ref{sec:bg}), we also aim for a system that 
relies only on data that can be obtained from relatively low-cost
instrumentation and public weather forecasts.
To achieve this, we rely on
recent developments in deep learning (presented in the second part
of Section~\ref{sec:bg}) and the adaptation of the machine learning
setup to the specifics of the task (Section~\ref{sec:core}).
We then present empirical results (Section~\ref{sec:exp})
and conclude (Section~\ref{sec:conc}).

\section{Background}
\label{sec:bg}

\subsection{Predicting thermal collector performance}

As already stated above, machine learning has been extensively
applied to construct models of solar thermal and PVT collectors.
\citet{ghritlahre-etal:2018} give a comprehensive review, and
show that these machine learning exercises are generally successful.
Solar radiation and ambient temperature serve are the key inputs in
most of these models, unsurprisingly, as physics-based models also
depend on these two variables.

What is worth noting about the relevant literature, is that it
is to a large extent addressing the design phase of collectors
or otherwise studying collectors outside the context of controlling
solar and auxiliary heating systems.
For instance, and moving forward to more recent works,
\citet{sadeghzadeh-etal:2019} compared
multi-layer perceptron, RBF, and Elman back-propagation ANNs
on the task of modelling the efficiency of solar collectors
at various flow rates of the working fluid; whereas
\citet{mirzaei-mohiabadi:2021} compared how the use of different working fluids affected the accuracy of ANN models
in predicting collector performance.
The study demonstrated the effectiveness of ANN in predicting
collector performance across all three fluids, highlighting
the potential of machine learning as a cost-effective and time-saving
alternative to conventional testing and modelling methods.

Closer to our work, \citet{gunasekar-etal:2015} used
Multi-Layered Perceptron Neural Networks to predict the thermal
performance of a PVT evaporator of a solar-assisted heat pump,
demonstrating close alignment with experimental data and
also identifying solar radiation and ambient temperature as the
primary input variables. \citet{du-etal:2022} trained a
convolutional neural network on a richer set of inputs
(solar radiation, ambient temperature, wind speed, fluid flow rate,
and fluid inlet temperature). The developed model showed strong
predictive capability in sunny conditions, but the authors
do not elaborate on the impact of the extended inputs on the 
prediction.
Additionally, \citeauthor{du-etal:2022} also note that
clustering analysis to screen the data and eliminate outliers
improves the accuracy of the prediction.

\subsection{Time representation}

One of the practical considerations often encountered in
timeseries processing is that fully observed, uniformly sampled
inputs are almost impossible to gather in realistic conditions.
Reasons include gaps in the data, varying sampling rates, and
(for multivariate timeseries) misalignment between variables' timing.

One approach for facing the reality of 
\textit{irregularly sampled and sparse multivariate timeseries}
is to first reconstruct a full, uniformly sampled timeseries
while another is to directly look for the underlying structure of
the data that is available. Reconstruction methods range from
simple imputation and aggregation \cite{marlin_etal, lipton_etal}
to sophisticated interpolation methods
\citep{yoon_2017_etal,shukla_marlin}.

Other works adjust the structure of known recurrent neural networks to directly handle irregularly sampled timeseries as input. \citet{che_etal} present several methods based on Gated Recurrent Units (GRUs), \citet{pham_etal} propose to capture time irregularity by modifying the forget gate of an LSTM and \citet{neil_etal} introduce a new time gate to be utilized within an LSTM. 
Another, more recent algorithm is proposed by \citet{mona_etal}
assuming a hidden state that evolves according to a linear
stochastic differential equation and is integrated
into an encoder-decoder framework.
\citet{rubanova_etal} present the ODE-RNNs, another auto-encode architecture, whose hidden state dynamics are specified by neural ordinary differential equations. 
These methods, however, fail to learn directly from partially observed vectors, which often occur when dealing with multiple variables.

Some early works \cite{li_etal, lu_etal} utilize the idea of Gaussian Processes (GPs) to model irregular timeseries. 
These works first optimize GP parameters and then train the corresponding model. 
\citet{futoma_etal} achieve end-to-end training for multivariate timeseries by using the re-parametrization trick to back-propagate the gradients through a black-box classifier. 
Though GPs provide a systematic way to deal with uncertainty, they are expensive to learn and are highly dependent on the chosen mean and covariance functions.

To go past these challenges, many researchers have also explored the use of Transformers. Unlike recurrent and differential equation-based architectures, which process inputs sequentially, Transformers use self-attention mechanisms \cite{og_attention} to capture relationships between all input positions simultaneously. Although they can be more efficient due to their parallel processing nature, they lack an inherent mechanism for representing temporal order. 
To address this limitation, a proposed method is to augment the input features with \term{time embeddings}.
When the data is known or suspected to exhibit periodicity,
\citet{wen-etal:2023} propose using sinusoidal functions that
separate periodicity and phase. For instance, adding as features
the sine and cosine of (numerical representation of)
months, days, hours as input features provides the model with
the means to capture periodicity at across different time scales. We shall refer to this model as CycTime.

Naturally, this representation is static and assumes prior knowledge
of which periodicities make sense for the phenomenon being modelled.
As expected, a line of research was developed to enhance the capabilities of attention-based architectures to inherently address the lack of time information. The multi-time attention network (mTAN) \cite{mTAN}, is similar to kernel-based interpolation with the difference that the attention-based similarity kernel is learnt. This model \emph{acquires} the time embeddings from a shallow
network that learns both \emph{periodic time patterns} and
\emph{non-periodic trends}.
These embeddings are then utilized by a multi-head attention mechanism that produces an embedding module composed of a set of reference time points and also re-represents the input features in a fixed-dimensional space.

This discussion motivates our choice of candidate models:
CycTime can take advantage of our prior knowledge of the existence of annual and daily periodicites in solar heat production, while mTAN can discover the interaction between the annual/daily periodicity
and the efficiency degradation trend of the collectors.

\section{Optimal Control for PVT Installations}
\label{sec:core}

As mentioned in the Introduction, our application targets the
mass-deployment of advanced control strategies in small domestic
installations. The high-level description of such an installation
and its operation is as follows:
\begin{itemize}
\item Hot water is produced by solar thermal or by
  PVT collectors that simultaneously produce
  thermal and electrical energy. Electricity production in the
  PVT case is not taken into account for the purposes of the work
  described here.
\item Hot water is used for both space heating and to cover domestic
  hot water demand. Ideally, only water from the solar thermal collector
  will be used for both usages, but in order to ensure meeting
  demand the system is also equipped with (less efficient) auxiliary
  heating systems.
\item The system decides automatically how to distribute the
  available hot water between space heating and domestic hot water
  demand and when to use the auxiliary heating system.
\end{itemize}

Our experimental installation is instrumented with
a flow rate meter, as well as inlet and outlet temperature sensors from
which hot water production can be
quantified.\footnote{These sensors are currently installed for
  our experimental purposes, but similar information can be
  obtained from low-cost domestic heat meters.}
Prior work has used this experimental installation on NCSR \quotes{Demokritos}
premises to test and validate this configuration for both office
(space heating only) and domestic (space heating and hot water) usage
profiles \citep{gg-etal:2022a,gg-etal:2022b}. The system is controlled
by static summertime and wintertime rules that are tailored to the local
climate. In our domestic application, the crucial quantity for these
rules is how the expected \term{heat production (\qpvt)} compares to
the expected \term{domestic hot water (DHW)} demand.

\qpvt can be directly calculated from sensor data as a function of
the input and output temperatures of the collector and the water flow.
Note that although some parameters are hardware dependent, they are
constant throughout an installation's life-cycle and do not need to be
re-evaluated.\footnote{Assuming the mirrors are regularly cleaned.}
But it can also be measured through the (experimentally determined)
\term{collector efficiency curve} that maps external conditions to \qpvt.
Using the collector efficiency curve has the advantage
that we can use weather forecasts to \emph{predict} the expected
\qpvt, but (as already noted) the curve changes as the collector ages
and needs to be re-evaluated.

What our application achieves is to \emph{continuously maintain} a
machine-learned surrogate of the collector efficiency curve by
using the installation's instrumentation to directly calculate \qpvt,
which is then used to label external conditions as provided by local
weather reports.

There is significant inherent uncertainty in the processes we
aim to model, thus it helps to abstract the prediction to bands of
\qpvt values that are meaningful for the given task.
Further to this, as the system
includes an insulated hot water tank, the timing of the production
does not need to closely match the timing of the demand, not even
for the morning spike as the previous day's water can be used,
if there was an excess.
In other words, we are not interested in the exact timing of the
production, but only if the system expects to run out of hot water
at some point during the day.

\begin{table}[bt]
\centering
\caption{\qpvt class thresholds (kWh)}
\label{tab:thresholds}
\begin{tabular}{ccc}
\hline
Balanced Ranges & Balanced Classes & Max Margins \\
\hline
  0.05          &  0.05            &   0.05      \\
  0.50          &  0.29            &   0.21      \\
  1.00          &  1.24            &   0.53      \\
  1.50          &  3.70            &   1.05      \\
\hline
\end{tabular}
\end{table}

This gives us the opportunity to reduce uncertainty by abstracting
the predictive task to wide bands of heat production over wide time
periods, instead of aiming to predict specific \qpvt values
at specific time-points. Since weather forecasts are published
for 3h periods, it makes sense to target this granularity in the
time dimension: Any finer predictions would simply repeat the
same inputs, or rely on interpolations that would add uncertainty;
any coarser predictions would have to aggregate the inputs
thus reducing the information available to the system.

Regarding the granularity at which the value of \qpvt is predicted,
we have tested three alternative approaches:
\begin{itemize}
\item Balanced ranges, simply separating the range of \qpvt values
  into bands of equal width.
\item Balanced classes, selecting thresholds that give a
  balanced dataset by simply sorting all values in the training data
  and cutting into equal sub-arrays.
\item Maximum margins, selecting thresholds that are as far as
  possible from expected levels of \emph{demand}, in order to
  offer downstream applications robustness to small divergence
  between expected and actual demand.
\end{itemize}
For the maximum-margins thresholds we referred to the
\term{DHW demand profiles} that are used to validate energy savings
from ecodesign of components \citep[Annex~III, Table~1]{eu_load_profiles}.
We set the thresholds to be exactly between the distinct values
appearing in that table. The specific threshold values are given in
Table~\ref{tab:thresholds}.\footnote{The maximum-margins thresholds
  in this table are based on the \term{medium profile}.
  It is straightforward to adjust them for different usage profiles.}

\section{Experiments and Results}
\label{sec:exp}

\begin{figure}[tb]
    \centering
    \includegraphics[scale=0.45]{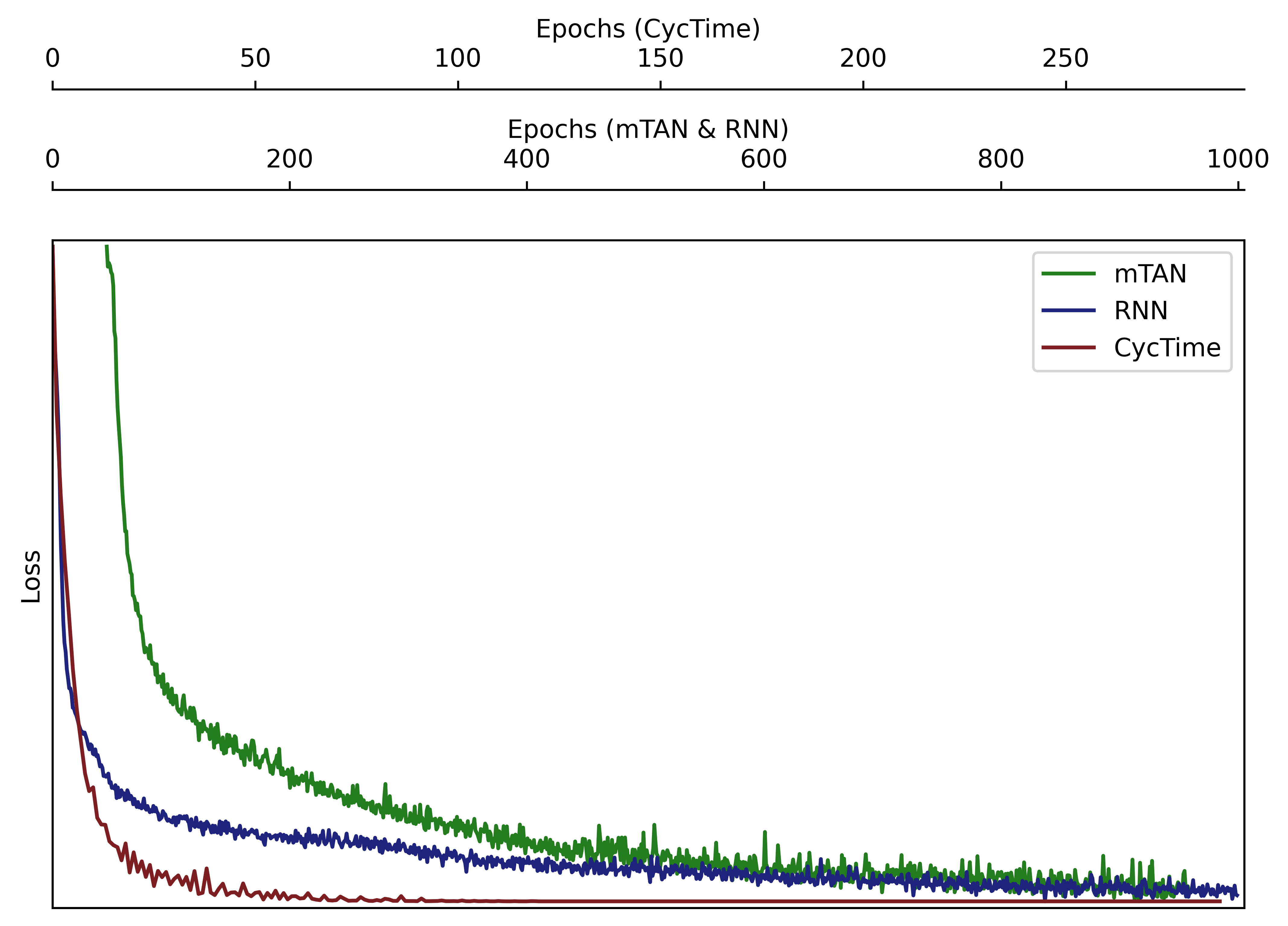}
    \caption{Normalized training loss progression for each model. Task is predicting the \quotes{Max Margins} bins. }
    \label{fig:train_losses}
\end{figure}


\begin{table*}[bt]
\centering
\caption{Evaluation results for the classification tasks.
  Each model was trained using six different initialization seeds, and the final metrics emerge from the aggregation (mean and standard
  deviation) of the results from these trained models.}

    \begin{tabular}{lccc}
        \toprule
        \multicolumn{4}{c}{\textbf{Balanced Ranges}} \\
        \midrule
        Metric & RNN & mTAN & CycTime  \\
        \midrule
        Precision (Micro) & $0.383 \pm 0.201$ & $0.740 \pm 0.032$ & $\mathbf{0.845} \pm 0.008$ \\
        Recall (Micro) & $ 0.383 \pm 0.201$ & $0.740 \pm 0.032$ & $\mathbf{0.845} \pm 0.008$ \\
        \rowcolor{blue!10} F-Score (Micro) & $ 0.383 \pm 0.201$ & $0.740 \pm 0.032$ & $\mathbf{0.845} \pm 0.008$ \\
        Precision (Macro) & $ 0.278 \pm 0.055$ & $0.580 \pm 0.028$ & $\mathbf{0.584} \pm 0.053$ \\
        Recall (Macro) & $ 0.298 \pm 0.027$ & $\mathbf{0.567} \pm 0.071$ & $0.553 \pm 0.020$ \\
        \rowcolor{blue!10} F-Score (Macro) & $0.237 \pm 0.060$ & $\mathbf{0.547} \pm 0.039$ & $ 0.544 \pm 0.026$ \\
        Precision (Weighted) & $ 0.532 \pm 0.102$ & $\mathbf{0.843} \pm 0.017$ & $\mathbf{0.843} \pm 0.018$ \\
        Recall (Weighted) & $ 0.383 \pm 0.201$ & $0.740 \pm 0.032$ & $\mathbf{0.845} \pm 0.008$ \\
       \rowcolor{blue!10}  F-Score (Weighted) & $ 0.372 \pm 0.192$ & $0.784 \pm 0.025$ & $\mathbf{0.833} \pm 0.014$ \\
        \toprule
        \multicolumn{4}{c}{\textbf{Balanced Classes}} \\
        \midrule
        Metric & RNN & mTAN & CycTime  \\
        \midrule
        Precision (Micro) & $ 0.395 \pm 0.221$ & $0.793 \pm 0.010$ & $\mathbf{0.794} \pm 0.008$ \\
        Recall (Micro) & $ 0.395 \pm 0.221$ & $0.793 \pm 0.010$ & $\mathbf{0.794} \pm 0.008$ \\
        \rowcolor{blue!10} F-Score (Micro) & $ 0.395 \pm 0.221$ & $0.793 \pm 0.010$ & $\mathbf{0.794} \pm 0.008$ \\
        Precision (Macro) & $0.332 \pm 0.120$ & $\mathbf{0.704} \pm 0.021$ & $0.653 \pm 0.013$ \\
        Recall (Macro) & $0.361 \pm 0.087$ & $\mathbf{0.760} \pm 0.020$ & $0.691 \pm 0.020$ \\
        \rowcolor{blue!10} F-Score (Macro) & $0.295 \pm 0.128$ & $\mathbf{0.718} \pm 0.021$ & $0.661 \pm 0.016$ \\
        Precision (Weighted) & $ 0.548 \pm 0.093$ & $\mathbf{0.825} \pm 0.013$ & $0.783 \pm 0.013$ \\
        Recall (Weighted) & $ 0.395 \pm 0.221$ & $0.793 \pm 0.010$ & $\mathbf{0.794} \pm 0.008$ \\
        \rowcolor{blue!10} F-Score (Weighted) & $0.392 \pm 0.203$ & $\mathbf{0.800} \pm 0.010$ & $0.784 \pm 0.010$ \\
        \toprule
        \multicolumn{4}{c}{\textbf{Max Margins}} \\
        \midrule
        Metric & RNN & mTAN & CycTime  \\
        \midrule
        Precision (Micro) & $ 0.453 \pm 0.143$ & $0.765 \pm 0.031$ & $\mathbf{0.831} \pm 0.009$ \\
        Recall (Micro) & $ 0.453 \pm 0.143$ & $0.765 \pm 0.031$ & $\mathbf{0.831} \pm 0.009$ \\
        \rowcolor{blue!10} F-Score (Micro) & $ 0.453 \pm 0.143$ & $0.765 \pm 0.031$ & $\mathbf{0.831} \pm 0.009$ \\
        Precision (Macro) & $0.310 \pm 0.041$ & $ 0.634 \pm 0.033$ & $\mathbf{0.684} \pm 0.037$ \\
        Recall (Macro) & $0.381 \pm 0.039$ & $\mathbf{0.701} \pm 0.037$ & $0.638 \pm 0.011$ \\
        \rowcolor{blue!10} F-Score (Macro) & $0.283 \pm 0.048$ & $\mathbf{0.651} \pm 0.037$ & $0.636 \pm 0.019$ \\
        Precision (Weighted) & $ 0.622 \pm 0.028$ & $\mathbf{0.822} \pm 0.016$ & $0.817 \pm 0.012$ \\
        Recall (Weighted) & $ 0.453 \pm 0.143$ & $0.765 \pm 0.031$ & $\mathbf{0.831} \pm 0.009$ \\
        \rowcolor{blue!10} F-Score (Weighted) & $0.447 \pm 0.153$ & $0.779 \pm 0.025$ & $\mathbf{0.813} \pm 0.008$ \\
        \bottomrule
    \end{tabular}

\label{tab:eval_scores}
\end{table*}

\subsection{Experimental Setup}

Our data was collected from the pilot building at NCSR \quotes{Demokritos}.
The building uses PVT for space heating and hot water, with a heat pump
as auxiliary heating system. The building is equipped with 
all necessary sensors to monitor the performance of the various
components tested there, including,
among others, a pyranometer (for solar radiation),
a temperature sensor with radiation shield (for ambient temperature),
temperature sensors for water/glycol inlet/outlet temperatures),
and flow rate meters (for water/glycol and hot water discharge).
Our focus for the work described here is on the part of the test facility
that includes 4 PVT collectors (total surface of 8.6 sq.m.) charging a
300-litre water tank. The medium tapping profile is programmed to remove
heat from this tank with the use of a controllable valve.

From the operation of the pilot
building we have acquired one year's worth of data, from
which \qpvt ground-truth can be calculated.
For the same period we have also obtained weather data i.e., temperature,
humidity levels, pressure, wind speed, accumulated rain and snow, measured
for the wider region that includes the installation.
We do not use weather forecasts, but actual weather data in order
to factor out the effect of weather forecasting uncertainty,
but the weather data we used is of the same kind and detail as the
forecasts published daily by the National Observatory of Athens.
As we mentioned already, weather forecasts are published for 3h periods,
so for our dataset we also aggregated \qpvt accordingly.

It is a common occurrence when collecting data from sensors to observe
missing values within the dataset due to defective sensors, networking
issues etc. In our case, we often have water temperature values
falling outside the accepted temperature range of
$-20^\circ\mathrm{C}$ to $100^\circ\mathrm{C}$. As \qpvt is calculated
from multiple water temperatures, it has approximately 15\% missing values
due to one or more of the variable needed to calculate \qpvt being missing.

Putting everything together, our machine learning task is to
predict a time-series of length 8 (because of the 3h aggregation)
where each prediction places \qpvt in one of the 5 classes
(value bins) defined by the thresholds in Table~\ref{tab:thresholds}.

\begin{figure*}[tb]
\centering
\includegraphics[scale=0.6]{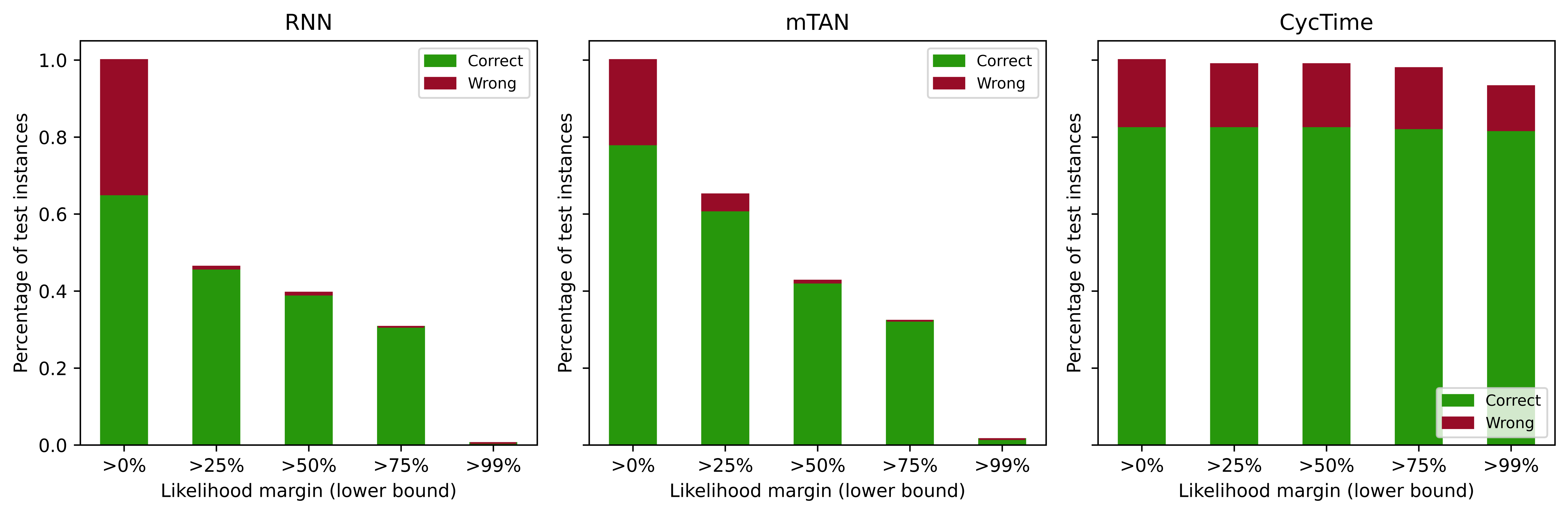}
\caption{The percentage of correct (green) and wrong (red) predictions,
split on lower bounds on the likelihood margin between the predicted
(most likely) class and the second most likely class.
Naturally, the left-most bar gives the accuracy on the complete test set.
Bars to the right give the accuracy on increasingly-confident decisions.}
    \label{fig:margin_errors}
\end{figure*}

\begin{figure}[tb]
\centering
\includegraphics[width=\linewidth]{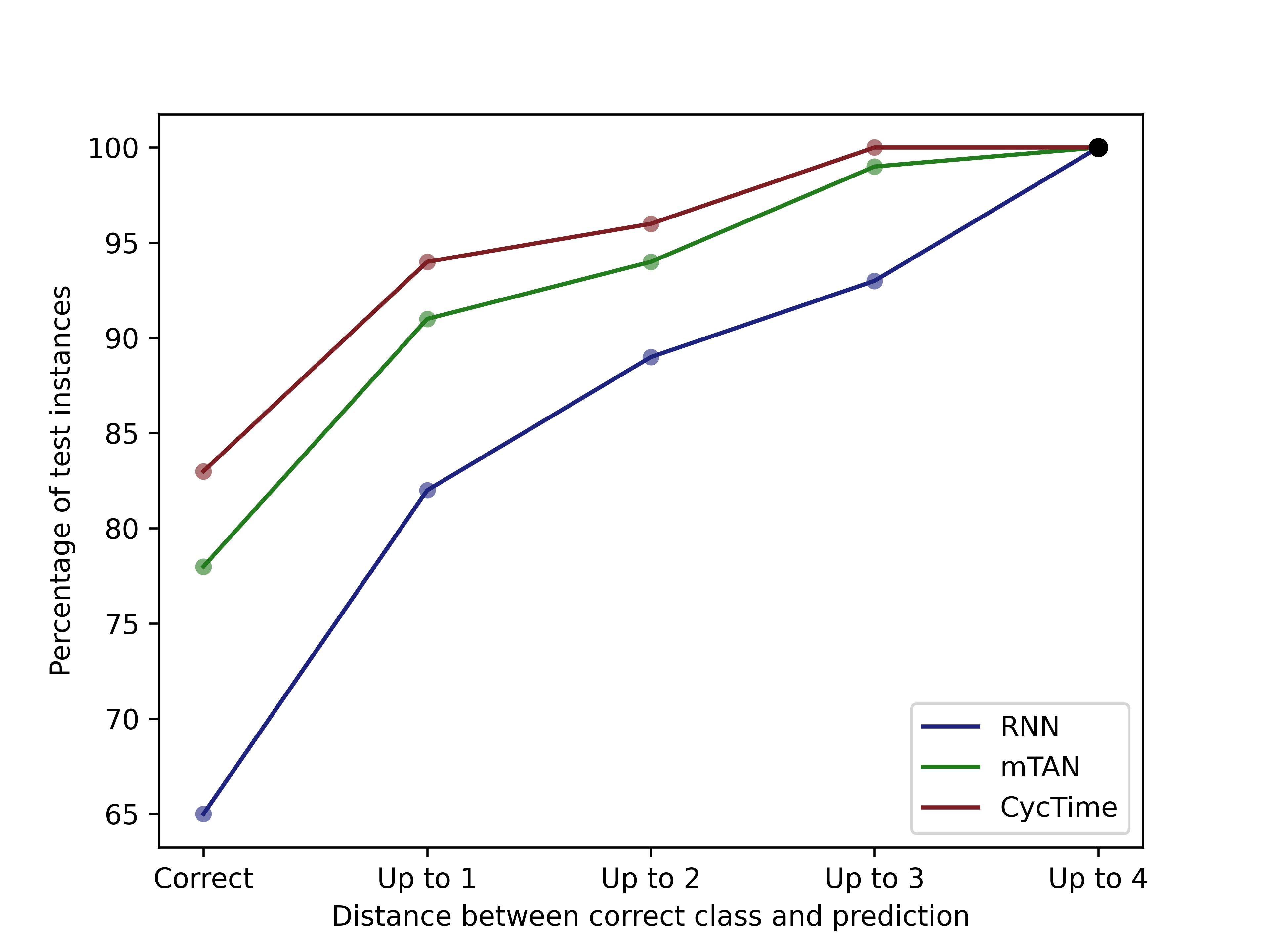}
\caption{The percentage of test instances that are predicted
of be of a class that is up to N classes away from the
ground-truth class, following the natural ordering of the
value ranges that each class represents.}
\label{fig:binning_results}
\end{figure}

\subsection{Model Architectures}

To perform the classification task we compare three methods: (a) a conventional RNN approach, (b) the CycTime transformer that uses fixed-time embeddings and (c) the mTAN module.

\paragraph{Baseline}
We include in out experiments a standard timeseries processing
approach as the baseline: linear interpolation to fill missing
and out-of-range sensor values, RNN, linear classification layer.

\paragraph{Transformer with fixed time embedding (CycTime)}
Our next approaches to tackle the classification problem are based on the idea of Transformers. This first method uses fixed positional time encodings. Regarding the model’s architecture, it consists of an encoder, a decoder, and a classifier. The encoder maps the input into a latent representation, while the decoder reconstructs it, preserving the original dimensionality of its features. The classifier layer then outputs a probability distribution over the five classes, with each element representing the probability of the input belonging to a particular class.

\paragraph{Multi-time attention network (mTAN)}
Adhering to the trajectory of attention-based approaches but employing a methodology based on learning the positional encodings of a timeseries, we apply the mTAN \cite{mTAN} module.
We deployed this module into an encoder-decoder architecture with the
decoder being a single linear classification layer, the same as the one used
in the previously discussed architectures.
We found that using 32 time-reference points achieved
better results. It is expected that the system should be given
some freedom to organize time in finer steps than the eight three-hour
steps it receives as input, and using a multiple of 8 makes it
straightforward to aggregate the outputs back to three-hour steps.
We found empirically that 32 steps and using mean as an aggregator
works best.\\

All models were trained using multi-class cross-entropy as the loss function, with an implemented mechanism to mask unobserved predicted labels.
Figure \ref{fig:train_losses} presents the models' training
loss progression over epochs.
Note that although the CycTime model appears to have the
longest training time per epoch (10 times longer than the mTAN model),
it requires 10 times fewer epochs on average to converge.
Consequently, both models require roughly the same overall training
time, which is 13-15 minutes on a single
NVIDIA RTX 5000 Ada Generation GPU. This level of computational load
should not pose any challenge for monthly or even weekly model updates.

\subsection{Results and Discussion}

We evaluated the performance of each model on the final two days of every month, data that we had excluded from the training set. 
The non-domain-specific evaluation metrics, which capture the micro, macro, and weighted recall, precision, and F-score achieved by the models in each predictive task, are presented in Table \ref{tab:eval_scores}. 
The micro average calculates the overall performance across all classes, while the macro average computes the average score across all classes, treating each class equally, and the weighted measurements calculate each metric with each class weighted by its support.
As anticipated, the RNN model exhibits poorer performance across all tasks,
while the other attention-based models compete closely with each other.

Regarding the choice of \qpvt value quantization, there does
not seem to be any dramatic advantage in class-balancing.
This affords us the flexibility to prefer the
max-margins quantization, which offers an advantage for
downstream processing. For this reason, we will continue the
discussion using results from the max-margins quantization.

Besides knowing which is the most accurate model, it is also
important to analyse their behaviour when they err. Since the task
is one of inherent uncertainty, we expect errors and we expect that
the downstream controller will be aware of the possibility of errors
and behave accordingly. This means that we are interested in the
graceful degradation of model accuracy so that the controller is not
completely thrown off and in having models that do not fail with
confidence, so that the control can trust high-confidence predictions.

\begin{table*}
\begin{center}
\caption{Percentage of all errors that occurred during
the early daylight period and during the main heat-producing
period.}
\label{tab:good_morning}
\vskip 2em

\begin{tabular}{lcccccccccc}
\toprule
        & \multicolumn{2}{c}{Balanced Ranges} &
          \multicolumn{2}{c}{Balanced Classes} &
          \multicolumn{2}{c}{Max Margins}  \\
\midrule    
        & \textbf{7am-10am} & \textbf{10am-7pm} &
         \textbf{7am-10am} & \textbf{10am-7pm} &
         \textbf{7am-10am} & \textbf{10am-7pm} \\
RNN         & 57\% & 41\% &
            54\% & 43\% &
            66\% & 34\% \\
mTAN        & 68\% & 32\% &
            69\% & 31\% &
            62\% & 31\% \\
CycTime        & 56\% & 44\% &
            58\% & 42\% &
            49\% & 51\% \\
\hline    
\end{tabular}
\end{center}
\end{table*}

\paragraph{Failing with confidence}

Figure \ref{fig:margin_errors} shows the confidence with
which the different models make predictions. 
Specifically, it displays the percentage of test instances that were classified correctly or misclassified with respect to different confidence margins of the prediction. 
Each prediction represents the distribution of likelihoods among five different value ranges. As a measure of confidence, we denote the likelihood difference between the best guess and the second best guess.
Ideally, we would like the likelihood mass for correctly predicted examples to be concentrated on a single output, while the likelihood mass for misclassified samples to be spread across the five different classes.
Simply put, model confidence is a positive quality only when its predictions are also accurate.
Although the RNN model exhibits the desired confidence behaviour, its accuracy is worse than that of the other two models.
The mTAN architecture has better accuracy than RNN, but with a similar behaviour when failing.
On the other hand the CycTime model, although more accurate than the other models, when it fails it fails with confidence.

\paragraph{Distance between true and predicted class}

Figure~\ref{fig:binning_results} shows the distance between the
predicted class and the ground-truth class for the different models.
There are no \quotes{reversals} of the relative ordering of the three
models, so as far as graceful degradation is concerned CycTime is
clearly preferred.

\paragraph{Morning errors}

Most errors occur between 7am and 10am, with the practically
all of rest falling between 10am and 7pm (cf. Table \ref{tab:good_morning}).
The high concentration of errors during the 7-10am period
can be attributed to the morning's dual nature, resembling
both day and night depending on the season. For instance, summer mornings are
sunny, resembling daytime, while winter mornings may still appear dark,
resembling night-time. It seems natural that CycTime is able to
handle this slightly better than the rest, since it is equipped
with the means to represent the annual cycle as priors.
We should remind at this point that we cannot change the
time aggregation in order to mitigate this, as it based on how
weather forecasts are published. 


\section{Conclusions and Future Work}
\label{sec:conc}

We have presented an application of the state-of-the-art in
Transformer models to the high-impact domain of renewable energy.
Specifically, we applied Transformer models to predict heat production
from solar collectors
to identify whether it is adequate to cover the demand and, ultimately,
to minimize the usage of (less efficient) auxiliary heating systems.

Based on the analysis presented above, we have decided to install
the mTAN model based on both theoretical and empirical reasons.
From the theoretical point of view, mTAN should be able to separate
the annual/daily cycles from the longer trend of collector efficiency
deterioration. We will revisit this empirically once the collectors
at our installation have aged enough to see perceptible changes,
but having this capability supports further applications such as
predictive maintenance.

From the empirical analysis, we have observed that having a static
prior time representation gives a noticeable but not impressive
advantage in prediction accuracy. However, mTAN has the advantage
that is less prone to failing with confidence: When it fails,
mTAN tends to have called a prediction on a slim difference in
likelihood, whereas the CycTime tends to have much
higher confidence in success as well as in failure.
mTAN's behaviour is more appropriate for our application, since
the controller can be configured to use the more conservative
or the more optimistic predictions, depending on the expected
demand and on the current condition of the system's hot water
tanks.

With the mTAN predictor installed and actively in use, we will
now be able to validate the actual energy gains by comparison
to the conventional (season-based) control previously used.
This will allow us to experiment with different policies
on how often to re-train the model and, in general, to validate
the complete concept of using installation's sensor data to
have the system automatically adapt.

In the analysis presented here we have used actual
weather conditions instead of weather forecasts, as the focus
was to see how well we can model the installation's internal
state. In future work, we will investigate the combined effect
of weather forecasting uncertainty and model uncertainty when
controlling conservatively or aggressively.
Finally, in further future work we plan to revisit the smart control
strategy with the aim to define a more dynamic strategy that
can \quotes{absorb} these uncertainties. Such a strategy would
exploit the system's internal state (e.g., the current charge of
the hot water tanks) and the knowledge of the margins of uncertainty
of future demand, weather forecasts, and the heat production predictor
to balance these uncertainties into an control that is optimal
in the long run.

\begin{ack}
This research has been co-financed by the European Union and
Greek national funds through the program
\quotes{Flagship actions in interdisciplinary scientific areas with
a special interest in the connection with the production network}
--- GREEN SHIPPING --- TAEDR-0534767 (Acronym: NAVGREEN).
For more information please visit \url{https://navgreen.gr}

\end{ack}

\end{document}